\documentclass{article}

\usepackage{arxiv}

\usepackage[utf8]{inputenc} 
\usepackage[T1]{fontenc}    
\usepackage{hyperref}       
\usepackage{url}            
\usepackage{booktabs}       
\usepackage{amsfonts}       
\usepackage{nicefrac}       
\usepackage{microtype}      
\usepackage{graphicx}
\usepackage{natbib}
\usepackage{subfigure}
\usepackage{graphicx}
\usepackage{caption}
\newcommand{\tabheader}[1]{%
  \bfseries
  \begin{tabular}{@{}c@{}}
  \strut#1\strut
  \end{tabular}%
}
\newcommand{\tabvalue}[1]{%
  \begin{tabular}{@{}l@{}}
  \strut#1\strut
  \end{tabular}%
}

\title{DeepRA: Predicting Joint Damage from Radiographs Using CNN with Attention}

\author{ Neelambuj Chaturvedi\\
	Advanced Data Science Group, 
	ZS Associates\\
	Bengaluru, India \\
	\texttt{neelambuj.nitj@gmail.com} 
	}

\date{}

\renewcommand{\headeright}{}
\renewcommand{\undertitle}{}

\hypersetup{
pdftitle={DeepRA}
}

\begin{document}
\maketitle

\begin{abstract}
Joint damage in Rheumatoid Arthritis (RA) is assessed by manually inspecting and grading radiographs of hands and feet. This is a tedious task which requires trained experts whose subjective assessment leads to low inter-rater agreement. An algorithm which can automatically predict the joint level damage in hands and feet can help optimize this process, which will eventually aid the doctors in better patient care and research. In this paper, we propose a two-staged approach which amalgamates object detection and convolution neural networks with attention which can efficiently and accurately predict the overall and joint level narrowing and erosion from patients radiographs. This approach has been evaluated on hands and feet radiographs of patients suffering from RA and has achieved a weighted root mean squared error (RMSE) of 1.358 and 1.404 in predicting joint level narrowing and erosion Sharp/van der Heijde (SvH) scores which is 31\% and 19\% improvement with respect to the baseline SvH scores, respectively. The proposed approach achieved a weighted absolute error of 1.456 in predicting the overall damage in hands and feet radiographs for the patients which is a 79\% improvement as compared to the baseline. Our method also provides an inherent capability to provide explanations for model predictions using attention weights, which is essential given the black box nature of deep learning models. The proposed approach was developed during the RA2 Dream Challenge hosted by Dream Challenges \footnote{http://dreamchallenges.org/challenges/} and secured 4\textsuperscript{th} and 8\textsuperscript{th} position in predicting overall and joint level narrowing and erosion SvH scores from radiographs.
\end{abstract}

\keywords{Attention \and CNNs \and Rheumatoid Artheritis \and Object Detection}

\section{Introduction}
Rheumatoid arthritis (RA) is an inflammatory rheumatic disease with progressive course affecting articular and extra-articular structures resulting in pain, disability, and mortality \cite{birch2010emerging}. Persistent inflammation leads to erosive joint damage and functional impairment in the vast majority of patients \cite{combe2009progression}. The extent of joint damage can be assessed from radiographs using the established and validated Sharp/van der Heijde (SvH) method \cite{van1995radiographic}. However, this requires experienced radiologists to manually inspect and grade the images. This is not only very time consuming, but also highly subjective, ultimately leading to low inter-rater reliability, where even trained experts often disagree on the final score \cite{renshaw2005comparison}. An automated scoring algorithm which can efficiently and accurately predict the narrowing and erosion SvH scores is an important and unmet need for better patient care for a number of reasons. First, RA-associated joint damage is not quantitative in clinical practice, as radiology reports typically state mild, moderate, or severe damage, which is highly subjective. Second, existing radiographs are a very objective measure of patient outcomes, but are not optimally used in patient care because of the lack of a method to quickly quantify damage. Third, there is often a disagreement between patient symptoms and the degree of damage in radiographs \cite{RA2_Dream_Challenge}. This discordance makes early identification and treatment of damage critical, as many biologics have clinical indications to prevent structural radiographic damage. Convolution Neural Networks (CNNs) have delivered outstanding performance in solving problems related to the medical imaging domain in the past, such as Classification of skin cancer \cite{fujisawa2019deep}, prediction of lung cancer mutations \cite{coudray2018classification}, prediction of covid 19 from chest X-Rays \cite{boudrioua2020covid} where the authors leverage deep neural networks to extract the hidden representation from the images and accurately predict the outcome and have achieved outstanding results. With recent developments in computer vision multitude of works have incorporated attention mechanism within CNNs which helps in boosting their representation learning capacity, such as CBAM, Dynamic convolution, Attention augmented CNNs \cite{woo2018cbam, bello2019attention, chen2020dynamic}. Attention mechanisms have also been implemented in the domain of medical imaging such as CABNet \cite{he2020cabnet} where the authors proposed an attention module which can be leveraged within CNNs for imbalanced diabetic retinopathy grading. Integrating attention blocks enhances the CNNs to focus on salient regions in the image but also acts as an inherent ability to explain the predictions made by the models. \par
In this paper, we propose a two staged approach for predicting joint narrowing and erosion SvH scores from hands and feet radiographs for patients using object detection and CNN with an attention mechanism. First stage being object detection, RetinaNet object detection models are trained on hands and feet radiographs to detect fingers/wrist joints from the radiographs. The second stage deals with developing joint level models which comprise of CNNs with attention to predict the joint-wise narrowing and erosion SvH scores on top of joints extracted from the first stage. Incorporating an attention mechanism assists the joint level models to focus on the salient regions in the joint radiographs to make damage predictions, additionally visualizing these attention maps helps in explaining the model predictions.
\par 
To summarize the contribution of this work
\begin{itemize}
\item[$\bullet$] We deal with the problem of non-availability of annotations for detecting fingers and wrist in hands and feet radiographs by training object detection models which can accurately detect these joints for further downstream tasks
\item [$\bullet$] We tackle the time consuming nature of grading radiographs by training joint level attention based CNN models which provide accurate predictions for narrowing and erosion SvH scores which also provides the overall RA damage for the patients
\item[$\bullet$] Given the black-box nature of deep learning models, our method provides inherent explanability using attention maps, which establishes confidence in the model predictions
\end{itemize}

Rest of the paper is organized as follows. In Section 2, we briefly describe the related works in this domain. We describe the overall methodology in detail i.e., object detection, CNN with attention in Section 3. In Section 4, we discuss the experiments and results and present the attention maps generated from the joint level models, we finally conclude the paper in Section 5.

\section{Related Work}
Earlier works on the radiographic scoring of patients with rheumatoid arthritis (RA) such as \cite{snekhalatha2017computer, langs2008automatic} focuses on solving the problem related to joint space narrowing, these methods lack extensive validation using the recently modified Sharp/van der Heijde (SvH) scores and these approaches also involve some degree of manual effort in generating hand crafted features. Other traditional methods to score radiographs are \cite{hoving2004comparison, dohn2008detection} which rely on magnetic resonance imaging  and sonography for damage predictions in patients with Rheumatoid arthritis. Some recent work which deals with scoring radiographs also exists for the degenerative, non-inflammatory osteoarthritis such as in \cite{tiulpin2018automatic} the authors proposed a deep siamese CNN for scoring the severity of osteoarthritis in knee radiographs and in \cite{xue2017preliminary} authors propose a method to detect osteoarthritis in hip radiographs using pretrained VGG \cite{simonyan2014very} models. More recently, the representation learning capabilities of CNNs have also been leveraged to categorize the level of joint damage from radiographs in limited works such as \cite{rohrbach2019bone} where the author developed a VGG \cite{simonyan2014very} inspired CNN model on hand radiographs at joint level to categorize the joint damage, another work which classifies the degree of joint damage from hand radiographs has been proposed in \cite{hirano2019development} where the authors use haar cascade features for joint detection and further leverage a CNN model to categorize the joint level damage, classifying the degree of joint damage can be higly subjective, additionally these approaches lack the inherent explainability of the model predictions. Other approaches developed during the challenge extracted joint patches from hands and feet radiographs using object detection and employed CNN models to predict the overall and joint level damage of the patients, although the performance of these approaches is promising however these approaches lacked inherent explanability for the model predictions. In this work, we propose a novel solution where we detect finger rather than finger joints from hands and feet radiographs and develop a attention-based CNN architecture which provides accurate predictions at par with the joint level approaches developed during the challenge and in prior works, additionally our approach has an inherent explainability for the model predictions.

\section{Methodology}
In this section, we discuss object detection using RetinaNet and Convolution Neural Networks with attention, which are leveraged in the proposed approach.

\subsection{Object Localization}
We address the challenge of localizing joints from radiographs using object detection, which exploits RetinaNet \cite{lin2017focal} a one-stage detector model which is able to accurately detect fingers/wrists from the radiographs. Object detectors are generally classified into two major categories i.e., two-stage detectors and one-stage detectors, where the former performs better in terms of accuracy in identifying the objects whereas the latter outperforms in terms of speed and simplicity. RetinaNet being a one-stage detector outperforms two-stage detectors in terms of accuracy, speed, and simplicity by utilizing a modified loss function known as focal loss \cite{lin2017focal} which tends to give higher weightage to hard examples and give a lower weight to easily classified examples, focal loss also reduces the contribution from easy examples and increases the importance of correcting misclassified examples. The architecture of RetinaNet constitutes of three major components, i.e., a backbone network for deep feature extraction, Feature Pyramid Network (FPN) to extract multi-scale features, and two subnetworks, i.e., one which predicts the class of the object and the second which predicts the coordinates of the bounding boxes in the input images. As shown In Fig. 1. RetinaNet employs backbone networks, such as residual networks (ResNet) \cite{he2016deep}, convolutional neural networks named by Visual Geometry Group (VGG) \cite{simonyan2014very}, or densely connected convolutional networks (DenseNet) \cite{huang2017densely} to extract higher semantic feature maps and then applies FPNs \cite{lin2017feature} to extract the multi-scale features from the backbone network, it adopts a pyramidal top-down architecture with lateral connections to capture the high-level feature maps at all scales and helps the model to detect objects of smaller size \cite{lin2017feature}. FPN extracts the features from the images at different scales and merges them with the features of consecutive layers of the backbone network, which effectively propagate these at different levels and scales to the succeeding layer \cite{li2019light}. The pyramidal features are then sent to the classification and regression subnets to classify and locate the desired objects.

\begin{figure}[h]
\centering
\includegraphics[width=14cm, height = 4.2cm]{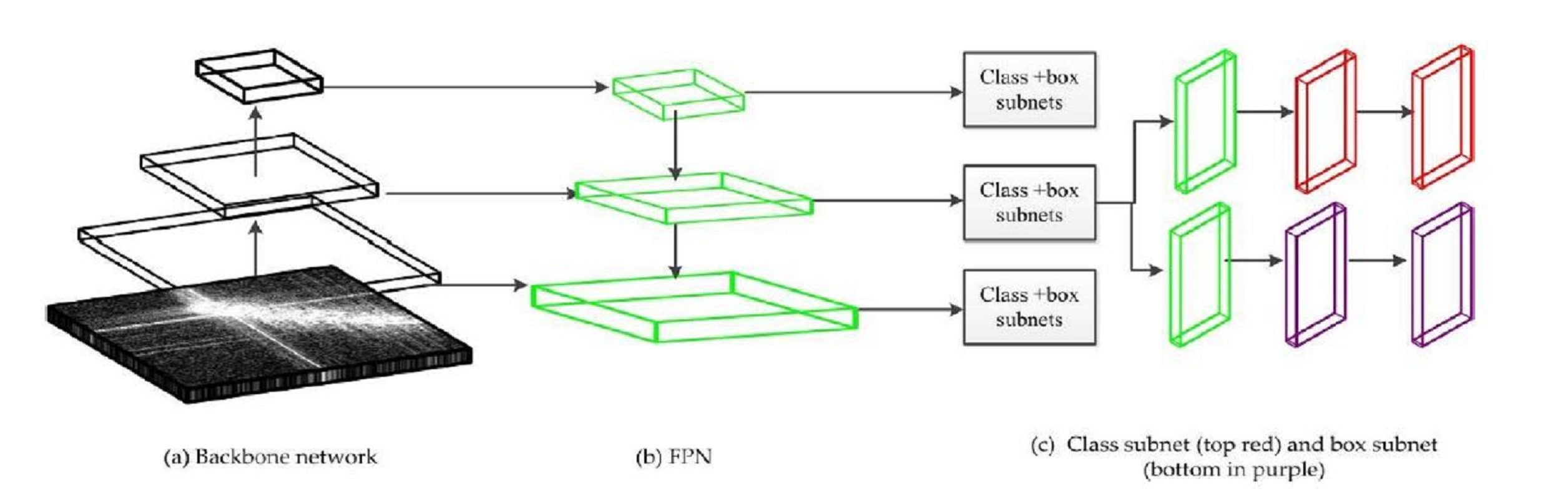}
    \caption{RetinaNet architecture consists of 3 major components a) Backbone Networks, b) Feature Pyramid Networks (FPN) and the c) Sub Networks. Backbone extracts the deep features from the image, FPN extracts the multi-scale features and merge these using pyramidal network structure. Finally, Sub Networks classifies and detects the objects in the image \cite{lin2014microsoft, wang2019automatic}}
\end{figure}

\begin{figure}[h]
    \centering
\includegraphics[width=15cm, height = 6cm]{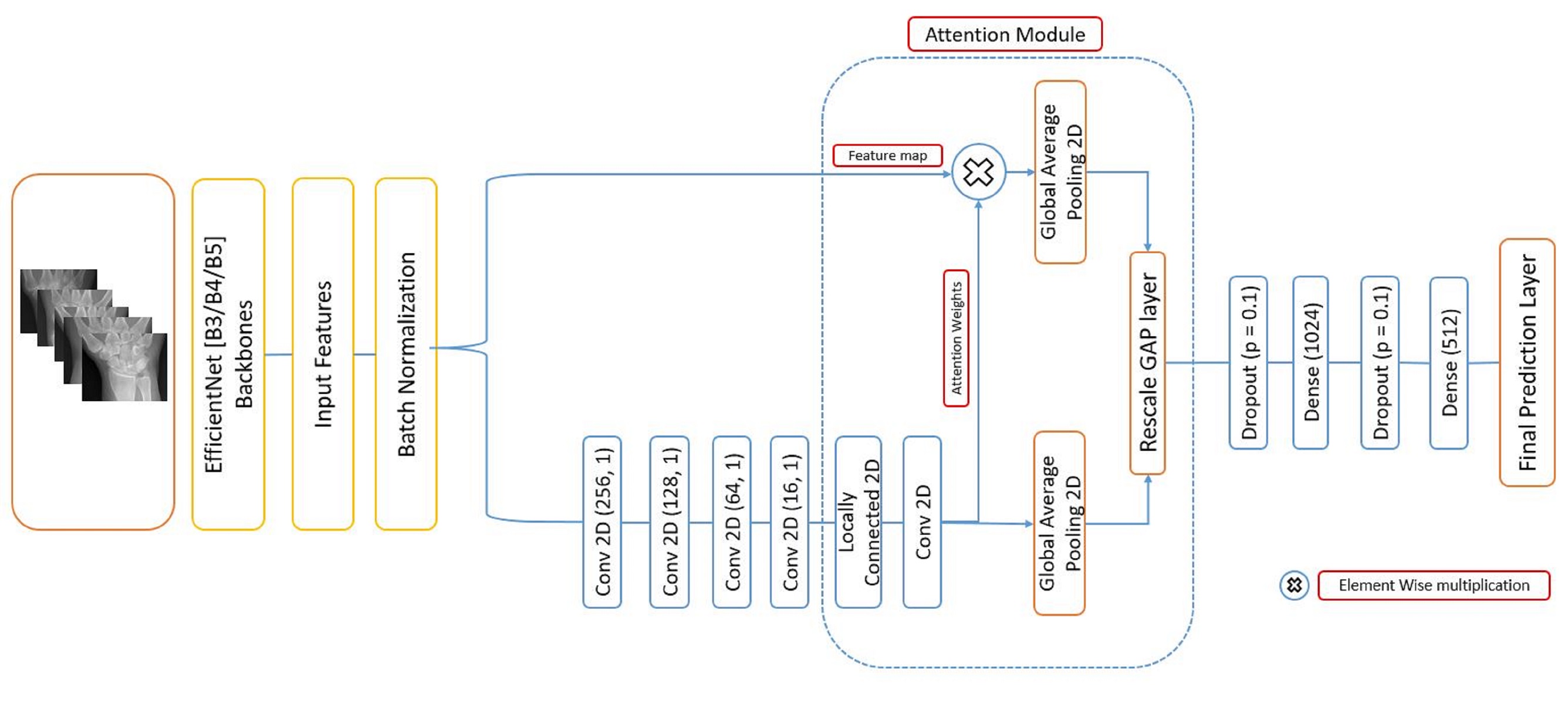}
    \caption{The Model architecture leverages EfficientNet B3-B5 \cite{tan2019efficientnet} backbones which are then passed on through the batch-normalization layer and further sub divided into two sub networks where one network convolves the feature maps and generate the attention weights which are then multiplied with second sub network element-wise and finally generate the narrowing/erosion scores after passing through the dense layers}
\end{figure}

\subsection{Convolution with Attention}
Convolutional Neural Network (CNN) has been recently employed to solve problems from both the computer vision and medical image analysis domains. Various research groups have developed innovative CNN architectures such as DenseNets \cite{huang2017densely}, ResNets \cite{he2016deep}, VGG \cite{simonyan2014very} which tend to be computationally expensive, Inception network utilizes a unique architecture which lowers the computational cost compared to the previous CNN architectures and achieves state-of-the-art performance in the ImageNet Large-Scale Visual Recognition Challenge 2014 (ILSVRC14) \cite{szegedy2015going}. In recent years, many attempts has been made to improve CNN performance by expanding them in different dimensions i.e width (W), depth (D) and resolution (R) with limited computational budget, although it is possible to scale two or three dimensions arbitrarily, this involves tedious manual tuning and generally yields substandard accuracy and efficiency. Increasing the width of the network helps the model to learn more complex features which causes a problem of vanishing gradients which has been mitigated with the use of skip connections as mentioned in \cite{he2016deep}. Wider network corresponds to the use of a higher number of channels which aim to capture more fine-grained patterns in the images. Using images with higher resolutions, \cite{huang2019gpipe} have also achieved state-of-the-art ImageNet accuracy. However, as we increase the W, D and R of the network, the accuracy quickly saturates after a certain point \cite{tan2019efficientnet}. Inspired by these observations more recently, a family of models named EfficientNets \cite{tan2019efficientnet} has been proposed, which tend to improve the performance of CNNs by scaling them in three dimensions, i.e., W, D, and R using a set of fixed scaling coefficients that meet a specific computational constraint. The family of EfficientNet models ranges from B0-B7 where B0 is the baseline network which was developed by neural architecture search (NAS) which tends to arrive at the optimal values for W, D, and R for a CNN given a computational constraint. Rescaling the architecture of the B0 model along three dimensions, i.e., W, D, and R constitute of 7 other models denoted by B1-B7. The EfficientNet-B1 is 7.6x smaller and 5.7x faster than ResNet-152 and EfficientNet-B7 achieves new state-of-the-art 84.3\% top-1 accuracy on ImageNet task but is 8.4x smaller and 6.1x faster than GPipe \cite{huang2019gpipe} as mentioned in \cite{tan2019efficientnet}. The proposed approach leveraged the EfficientNet B3-B5 models as backbones in the joint level models to grade radiographs.
\subsubsection{Attention} 
Attention mechanism has revolutionized the Natural Language Processing (NLP) domain. More recently numerous efforts have been made to incorporate attention mechanisms in CNNs such as \cite{xu2015show} where authors introduce the concept of soft and hard attention in images to generate image captions. We incorporate an attention mechanism in the EfficientNet architecture which enhances the representation learning capacity of the EfficientNet architecture. This proposed approach is inspired by \cite{schlemper2019attention} in which the authors apply attention gated CNN network for object localization in a 3D CT abdominal dataset. Attention module not only helps the model to eventually learn the salient regions in the radiographs while making predictions but also adds a layer of explainability by extraction of attention weights. These weights are obtained by passing the input feature maps from the EfficientNet backbone through various convolution layers and finally utilizing a locally connected 2D layer with sigmoid activation which computes the weights at a pixel level ranging from 0-1. A element-wise multiplication of these attention weights with the feature maps helps the model to turn off the pixel values where the weights are zero. Fig. 2. depicts the final model architecture which is trained on the extracted joints from the first stage, i.e., object detection. We employ task-specific EfficientNet backbones and train an end-to-end network which predicts the joint damage from the radiographs. Code for this approach is publicly available at https://github.com/NC717/DeepRA.

\section{Experimentation and Results}
\subsection{Dataset used}
The Dataset used, for the analyses in this paper was contributed by University of Alabama at Birmingham. This dataset was obtained as part of the RA2-DREAM Challenge: Automated Scoring of Radiographic Damage in Rheumatoid Arthritis through Synapse ID [syn20545111] \cite{bridges2010radiographic, ormseth2016effect}. The training dataset consisted of 1,468 radiographs corresponding to 367 patients with manually annotated narrowing and erosion SvH scores at a patient level. Leaderboard performance was evaluated on hands and feet radiographs for 300 patients which along-with the final evaluation test set was not available to the participants during the course of the RA2 Dream Challenge. SvH scoring system considers joints in the hands/wrist and feet that are typical targets of inflammation in rheumatoid Arthritis (RA), which includes joints in the wrists, metacarpal phalangeal (MCP) and proximal interphalangeal (PIP) joints of the hands, and the metatarsal phalangeal (MTP) and the PIP joints in feet. The challenge was to predict narrowing and erosion scores corresponding to these joints at a patient level, which can further be used to get the overall damage in the radiographs \cite{Sun2021.10.25.21265495}.



\subsection{Joint localization}
First stage of the proposed approach, i.e., object detection, was performed to extract the fingers and wrist from the radiographs for both hands and feet.
\subsubsection{Implementation Details}
We trained RetinaNet model with ResNet50 \cite{he2016deep} as the backbone network, which was pretrained on COCO image dataset \cite{lin2014microsoft}. All weights of the backbone network were trainable owing to the inherent differences in the data distributions in COCO dataset and the hands and feet radiographs. Fingers in hands and feet and wrist in hand were manually annotated for 914 radiographs using an open source tool labellmg\footnote{https://github.com/tzutalin/labelImg}, which were finally used to train and validate two separate detector models to detect fingers/wrist from the radiographs. First model was developed to detect fingers and wrist in hand and the second model detected fingers in feet. We had 341 and 57 radiographs in the training and validation sets, respectively, to develop the hand finger/wrist detection model, whereas the feet finger detection model consisted of 355 and 61 radiographs in the training and validation sets, respectively. During training, we limit the dimensions of the input radiographs with image Min/Max dimension bounded at 1,000 and 1,400 respectively. We employ Adam \cite{kingma2014adam} optimizer with a learning rate of 0.001 which is reduced during training by a factor of 0.1 if the mean Average Precision (mAP) has not improved after 4 consecutive epochs. Batch size of 8 is used during training and to reduce the overfitting, early stopping is used if the mAP has not improved for 5 consecutive epochs on the validation set. We use Tesla K80 GPU on Google Colab to train and test the models. 
\subsubsection{Results}
We evaluate the detector models on the test dataset which comprised of 50 Hands/Feet radiographs respectively, these radiographs were not used during the training and validation. The detector model, for the feet was able to achieve a mAP of 0.965, whereas the hand model achieved a mAP of 0.981 at an intersection over union threshold (IOU) of 0.6. In Table 1, we report the joint wise average precision (AP), we observe that the wrist joint being the easiest to detect, achieves the highest AP, whereas all other joints have a relatively similar AP in the case of both detector models. Fig. 4 represents the joints detected for a patient in the test set.
\begin{figure}[h]
    \centering
    \begin{subfigure}[ Hand Fingers/Wrist detected by the model]{{\includegraphics[height=7cm, width=6.8cm]{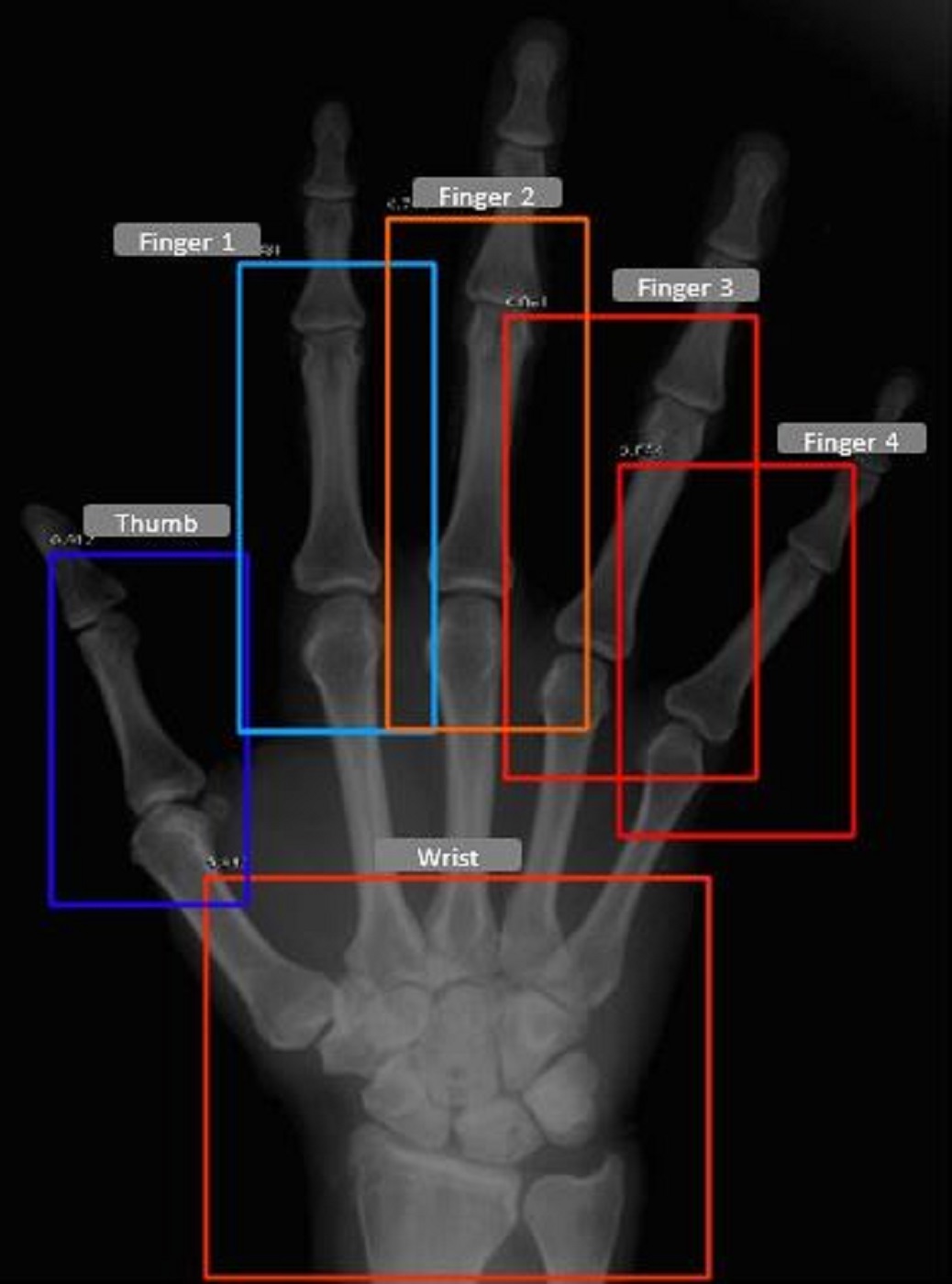}}}
    \qquad
    \end{subfigure} \hspace{1cm}
    \begin{subfigure}[ Feet Fingers/Toe detected by the model]{{\includegraphics[height=7cm, width=6.8cm]{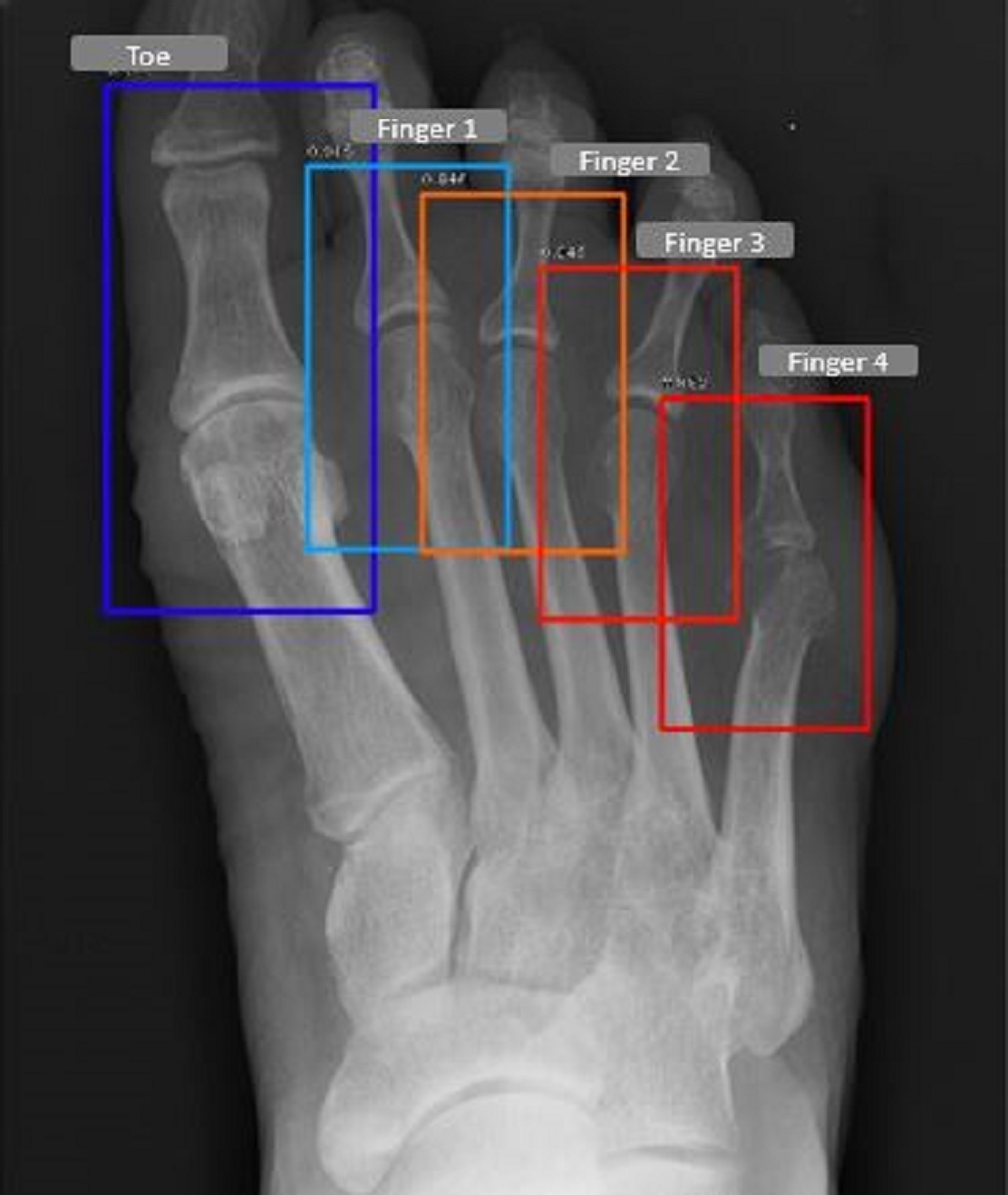} }}
    \caption{Fingers, wrists and thumb/toe identified by the detection models for a patient}
        \end{subfigure}%
\end{figure}

\begin{table}[h]
\renewcommand{\tabcolsep}{3mm}
\caption{Overall and joint wise Average Precision (AP) at IOU threshold of 0.6}\label{tab3}
\centering
\begin{tabular}{| l |  c   c |}
\specialrule{.1em}{.05em}{.05em}
\tabheader{Joint Type} & \tabheader{Hand Model \\ Joint Wise AP} & \tabheader{Feet Model \\ Joint Wise AP}\\
\hline
Thumb/Toe & 0.986  & 0.983 \\
\hline
Finger 1 &  0.979 & 0.973 \\
\hline
Finger 2 &  0.983 & 0.955 \\
\hline
Finger 3 &  0.971 & 0.947 \\
\hline
Finger 4 &  0.977 & 0.965 \\
\hline
Wrist &  0.993 & - \\
\hline
\textbf{Overall mAP} &  \textbf{0.981} & \textbf{0.965} \\
\specialrule{.1em}{.05em}{.05em}
\end{tabular}
\end{table}

\subsection{Joint Level Models}
Second stage of the proposed approach consists of training joint level models, to predict the joint narrowing and erosion SvH scores from the radiographs. We describe the implementation details and the results in the following sections.

\subsubsection{Implementation Details}
We train 6 joint level models which are based on the EfficientNet [B3-B4-B5] backbones pretrained on ImageNet dataset along-with the attention mechanism. Finger joints extracted from object detection stage were resized to $256\times 256$ whereas wrist joints were resized to $300\times300$ while training. We use a 80:20 train test split to train all joint level models. During training, we use Adam optimizer \cite{kingma2014adam} and mean squared error (MSE) as the loss function with an initial learning rate of 0.001 which is decayed by a factor of 0.2 if the validation RMSE does not improve within 5 epochs. We train the models for 200 epochs and finally save the model which has the least RMSE on the validation set. To make the model robust, we employ horizontal flipping of the joint images in every joint model. During training 70\% weights of the EfficientNet backbones were unfreezed since the distribution of the medical radiographs is entirely different from the ImageNet dataset which was originally used to train these models. Approximately 70\% of the joints had a Narrowing/Erosion SvH score of zero in the training data which caused the issue of imbalance while training. We tackled this by training the model by randomly sampling the equivalent number of joints with non-zero Narrowing/Erosion scores from zero damage joint radiographs, so that there is no class imbalance while training the model. We use a batch-size of 32 and all the models are implemented with Keras using Tensorflow backend. All the experiments are performed on Tesla K80 GPUs on Google Colab. In Table 2, we describe the training specifications for the joint level models which describe the EfficientNet backbones utilized, the number of joint images extracted from the radiographs using the detector models, and the number of predictions to be made in every model.
 $$ MSE = (\frac{1}{n})\sum_{i=1}^{n}(y_{i} - x_{i})^{2} $$

\begin{table}[h]
\renewcommand{\tabcolsep}{2mm}
\centering
\caption{\textbf{Training specifications for Joint level models:} This table represents the EfficientNet backbones used, the number of joints extracted from object detection stage and number of predictions to be made for every joint level model}\label{tab2}
\begin{tabular}{| l |  l |  c  c  c| }
\specialrule{.1em}{.05em}{.05em}
\textbf{Model type} &  \tabheader{Finger Joints} & \tabheader{EfficientNet \\Backbone} &  \tabheader{Total Joint \\Images} &  \tabheader{Number of\\Predictions}\\
\hline
Erosion & Hand Fingers \& Thumb/Toe & B3  & 4,404 & 2\\
\hline
Narrowing & Hand Fingers \& Toe &  B3  & 3,670 & 2 \\
\hline
Narrowing  & Feet Fingers \& Thumb &  B4 & 3,670 & 1\\
\hline
Erosion & Feet Fingers &  B4   & 2,936 & 1\\
\hline
Erosion \& Narrowing & Wrist &  B5   & 734 & 6\\
\specialrule{.1em}{.05em}{.05em}
\end{tabular}
\end{table}

\subsubsection{Evaluation Metrics}
The RA2 Dream challenge utilized the weighted (Wt.) root mean squared error (RMSE) to evaluate the joint level narrowing and erosion SvH scores, whereas weighted (Wt.) absolute error was used to evaluate the overall RA damage at a patient level. The weights used for calculating the joint wise RMSE were not released during the RA2 Dream Challenge. 
 $$ Weighted \hspace{1.5mm} RMSE = \sqrt{(\frac{1}{n})\sum_{i=1}^{n}(w_{i})(y_{i} - x_{i})^{2}} $$
 $$ Weighted \hspace{1.5mm} MAE = (\frac{1}{n})\sum_{i=1}^{n}(w_{i})\left | y_{i} - x_{i} \right | $$


\subsubsection{Results}
The proposed approach was evaluated on leaderboard and the final test dataset which comprised on 1200 radiographs corresponding to 300 patients each, these datasets were not available to the participants during the challenge. The RA2 Dream baseline uses the distribution of scores in the training data to create a random prediction. In Table 3, we describe multiple experiments and the associated Weighted AE for overall RA damage prediction and Weighted RMSE for joint narrowing and erosion SvH score prediction on the Leaderboard/Final test datasets. Initial experiments were performed with a DenseNet-169 model pretrained on ImageNet dataset which is fine tuned further on hands and feet radiographs without joint detection and attention to predict the joint wise and overall SvH scores, which showed a minor improvement w.r.t baseline score, whereas when we include joint detection and attention into the same model architecture we were able to improve the model performance significantly. We further went on and incorporated various pretrained EfficientNet backbones into the joint level model architecture and observed that the EfficientNet models with joint detection and attention outperformed all the other experiments which were executed during the challenge. Our EfficientNet based joint models with Joint detection and attention achieved 4\textsuperscript{th} place in predicting the overall damage and 8\textsuperscript{th} place in predicting the joint Narrowing and Erosion SvH scores on the final test dataset.

\begin{table}
\renewcommand{\tabcolsep}{2mm}
\caption{\textbf{Model performance on the public/private leaderboard: } EfficientNets with attention outperforms other models in joint-wise narrowing prediction and overall damage prediction from the radiographs}
\label{tab1}
\centering
\begin{tabular}{| l |  c  c  c|  }
\specialrule{.1em}{.05em}{.05em}
\textbf{Experiments} & \tabheader{Weighted Abs. Error \\(Overall)} & \tabheader{ Weighted RMSE \\(Narrowing)} & \tabheader{Weighted RMSE \\(Erosion)} \\
\hline
Baseline & 7.023  & 1.96 & 1.728\\
\hline
\tabvalue{DenseNet-169 w/o Joint detection \& Attention} &  2.979 & 1.659  & 1.487 \\
\hline
\tabvalue{DenseNet-169 w. Joint Detection \& Attention} &  1.465 & 1.449  & 1.394 \\
\hline
\tabvalue{EfficientNets w. Joint Detection \& Attention} &  1.371 & 1.433  & 1.397\\
\hline
\tabvalue{Final Performance (EfficientNet \& Attention)} &  \textbf{1.456} & \textbf{1.358}  & \textbf{1.404}\\
\specialrule{.1em}{.05em}{.05em}
\end{tabular}
\end{table}

\subsection{Attention Map Analysis}
While dealing with healthcare related problems we need to be sure of what we are predicting for a patient. In this case since we are predicting joint level narrowing and erosion SvH scores we need to have some evidence in our model predictions, otherwise it is tough to acknowledge that the predictions are accurate. Here we demonstrate that the attention mechanism we introduced within the model architecture is not only capable of boosting the model performance but also provides a layer of explanability for the model predictions. We extracted the attention weights and overlayed them onto the joint images. In Fig. 5, we demonstrate the attention maps for narrowing SvH score prediction problem for hand/feet fingers, and wrist. 
\\
\begin{figure}[h]
    \centering
    \begin{subfigure}[\centering Feet finger]{{\includegraphics[height=4.8cm, width=5cm]{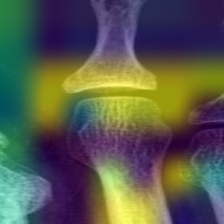} }}
    \end{subfigure}
    \begin{subfigure}[\centering Hand finger ]{{\includegraphics[height=4.8cm, width=5cm]{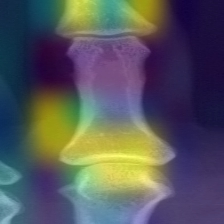} }}
    \end{subfigure}
    \begin{subfigure}[\centering Wrist]{{\includegraphics[height=4.8cm, width=5cm]{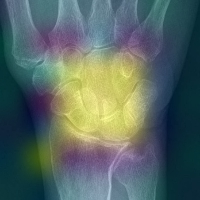} }}
    \end{subfigure}
    \caption{Attention weights corresponding to various joints while predicting narrowing scores from extracted fingers and wrists from the radiographs, yellow regions demonstrate the region where the model assigns higher weights in the input image}
\end{figure}
\\
In Fig. 5(a), and 5(b), the joint region in the feet and hand fingers is activated to provide the final predictions, similarly, we see that in Fig. 5(c), all wrist joints are given a higher weightage by the model while making the predictions. Our attention map provides an inbuilt explanability component which can aid the radiologist to believe in the model predictions.
\section{Conclusion}
In this paper, we propose an automated method for detecting joints from radiographs and predicting the associated damage in the joints by leveraging attention based CNN framework. Due to the lack of availability of annotations for the the joints we propose an object detection model which is capable of correctly identifying and localizing the joints in the radiographs which can be leveraged for downstream tasks. Since manual grading of radiographs requires experienced radiologists, we tackle this with an attention-based CNN method which can accurately grade the radiographs and aid radiologists in better patient care and research. Not only the model is able to give better predictions but it also provides a sense of the salient regions in the joint radiographs where the model latches on to make those predictions. While the achieved performance is encouraging, further evaluation of this approach on broader dataset is required to have a more reliable estimation of the model predictions. In the future, we aim to develop joint level detector models instead of a finger level models and employ the similar model architecture which could provide a great deal of understanding while predicting the scores and assist the radiologists in making their decisions, also directly comparing the explainability of inherent attention maps and other offline explainability methods may instill confidence in the predictions.

\bibliographystyle{unsrtnat}
\bibliography{references}

\begin{thebibliography}{37}
\providecommand{\natexlab}[1]{#1}
\providecommand{\url}[1]{\texttt{#1}}
\expandafter\ifx\csname urlstyle\endcsname\relax
  \providecommand{\doi}[1]{doi: #1}\else
  \providecommand{\doi}{doi: \begingroup \urlstyle{rm}\Url}\fi

\bibitem[Birch and Bhattacharya(2010)]{birch2010emerging}
James~T Birch and Shelley Bhattacharya.
\newblock Emerging trends in diagnosis and treatment of rheumatoid arthritis.
\newblock \emph{Primary Care: Clinics in Office Practice}, 37\penalty0
  (4):\penalty0 779--792, 2010.

\bibitem[Combe(2009)]{combe2009progression}
Bernard Combe.
\newblock Progression in early rheumatoid arthritis.
\newblock \emph{Best Practice \& Research Clinical Rheumatology}, 23\penalty0
  (1):\penalty0 59--69, 2009.

\bibitem[Van~der Heijde et~al.(1995)Van~der Heijde, Van~Leeuwen, Van~Riel, and
  Van~de Putte]{van1995radiographic}
DMFM Van~der Heijde, MA~Van~Leeuwen, PLCM Van~Riel, and LBA Van~de Putte.
\newblock Radiographic progression on radiographs of hands and feet during the
  first 3 years of rheumatoid arthritis measured according to sharp's method
  (van der heijde modification).
\newblock 1995.

\bibitem[Renshaw and Gould(2005)]{renshaw2005comparison}
Andrew~A Renshaw and Edwin~W Gould.
\newblock Comparison of disagreement and error rates for three types of
  interdepartmental consultations.
\newblock \emph{American journal of clinical pathology}, 124\penalty0
  (6):\penalty0 878--882, 2005.

\bibitem[RA2(2019 (accessed January, 2021))]{RA2_Dream_Challenge}
Ra2 dream challenge.
\newblock \url{https://www.synapse.org/#!Synapse:syn20545111/wiki/597242}, 2019
  (accessed January, 2021).

\bibitem[Fujisawa et~al.(2019)Fujisawa, Otomo, Ogata, Nakamura, Fujita,
  Ishitsuka, Watanabe, Okiyama, Ohara, and Fujimoto]{fujisawa2019deep}
Y~Fujisawa, Y~Otomo, Y~Ogata, Y~Nakamura, R~Fujita, Y~Ishitsuka, R~Watanabe,
  N~Okiyama, K~Ohara, and M~Fujimoto.
\newblock Deep-learning-based, computer-aided classifier developed with a small
  dataset of clinical images surpasses board-certified dermatologists in skin
  tumour diagnosis.
\newblock \emph{British Journal of Dermatology}, 180\penalty0 (2):\penalty0
  373--381, 2019.

\bibitem[Coudray et~al.(2018)Coudray, Ocampo, Sakellaropoulos, Narula, Snuderl,
  Feny{\"o}, Moreira, Razavian, and Tsirigos]{coudray2018classification}
Nicolas Coudray, Paolo~Santiago Ocampo, Theodore Sakellaropoulos, Navneet
  Narula, Matija Snuderl, David Feny{\"o}, Andre~L Moreira, Narges Razavian,
  and Aristotelis Tsirigos.
\newblock Classification and mutation prediction from non--small cell lung
  cancer histopathology images using deep learning.
\newblock \emph{Nature medicine}, 24\penalty0 (10):\penalty0 1559--1567, 2018.

\bibitem[Boudrioua(2020)]{boudrioua2020covid}
Mohamed~Samir Boudrioua.
\newblock Covid-19 detection from chest x-ray images using cnns models: Further
  evidence from deep transfer learning.
\newblock \emph{Available at SSRN 3630150}, 2020.

\bibitem[Woo et~al.(2018)Woo, Park, Lee, and So~Kweon]{woo2018cbam}
Sanghyun Woo, Jongchan Park, Joon-Young Lee, and In~So~Kweon.
\newblock Cbam: Convolutional block attention module.
\newblock In \emph{Proceedings of the European conference on computer vision
  (ECCV)}, pages 3--19, 2018.

\bibitem[Bello et~al.(2019)Bello, Zoph, Vaswani, Shlens, and
  Le]{bello2019attention}
Irwan Bello, Barret Zoph, Ashish Vaswani, Jonathon Shlens, and Quoc~V Le.
\newblock Attention augmented convolutional networks.
\newblock In \emph{Proceedings of the IEEE International Conference on Computer
  Vision}, pages 3286--3295, 2019.

\bibitem[Chen et~al.(2020)Chen, Dai, Liu, Chen, Yuan, and Liu]{chen2020dynamic}
Yinpeng Chen, Xiyang Dai, Mengchen Liu, Dongdong Chen, Lu~Yuan, and Zicheng
  Liu.
\newblock Dynamic convolution: Attention over convolution kernels.
\newblock In \emph{Proceedings of the IEEE/CVF Conference on Computer Vision
  and Pattern Recognition}, pages 11030--11039, 2020.

\bibitem[He et~al.(2020)He, Li, Li, Wang, and Fu]{he2020cabnet}
Along He, Tao Li, Ning Li, Kai Wang, and Huazhu Fu.
\newblock Cabnet: Category attention block for imbalanced diabetic retinopathy
  grading.
\newblock \emph{IEEE Transactions on Medical Imaging}, 2020.

\bibitem[Snekhalatha and Anburajan(2017)]{snekhalatha2017computer}
U~Snekhalatha and M~Anburajan.
\newblock Computer-based measurements of joint space analysis using metacarpal
  morphometry in hand radiograph for evaluation of rheumatoid arthritis.
\newblock \emph{International journal of rheumatic diseases}, 20\penalty0
  (9):\penalty0 1120--1131, 2017.

\bibitem[Langs et~al.(2008)Langs, Peloschek, Bischof, and
  Kainberger]{langs2008automatic}
Georg Langs, Philipp Peloschek, Horst Bischof, and Franz Kainberger.
\newblock Automatic quantification of joint space narrowing and erosions in
  rheumatoid arthritis.
\newblock \emph{IEEE transactions on medical imaging}, 28\penalty0
  (1):\penalty0 151--164, 2008.

\bibitem[Hoving et~al.(2004)Hoving, Buchbinder, Hall, Lawler, Coombs, McNealy,
  Bird, and Connell]{hoving2004comparison}
Jan~Lucas Hoving, Rachelle Buchbinder, Stephen Hall, Gary Lawler, Peter Coombs,
  Stephen McNealy, Paul Bird, and David Connell.
\newblock A comparison of magnetic resonance imaging, sonography, and
  radiography of the hand in patients with early rheumatoid arthritis.
\newblock \emph{The Journal of Rheumatology}, 31\penalty0 (4):\penalty0
  663--675, 2004.

\bibitem[D{\o}hn et~al.(2008)D{\o}hn, Ejbjerg, Hasselquist, Narvestad,
  M{\o}ller, Thomsen, and {\O}stergaard]{dohn2008detection}
Uffe~M{\o}ller D{\o}hn, Bo~J Ejbjerg, Maria Hasselquist, Eva Narvestad, Jakob
  M{\o}ller, Henrik~S Thomsen, and Mikkel {\O}stergaard.
\newblock Detection of bone erosions in rheumatoid arthritis wrist joints with
  magnetic resonance imaging, computed tomography and radiography.
\newblock \emph{Arthritis research \& therapy}, 10\penalty0 (1):\penalty0 R25,
  2008.

\bibitem[Tiulpin et~al.(2018)Tiulpin, Thevenot, Rahtu, Lehenkari, and
  Saarakkala]{tiulpin2018automatic}
Aleksei Tiulpin, J{\'e}r{\^o}me Thevenot, Esa Rahtu, Petri Lehenkari, and Simo
  Saarakkala.
\newblock Automatic knee osteoarthritis diagnosis from plain radiographs: a
  deep learning-based approach.
\newblock \emph{Scientific reports}, 8\penalty0 (1):\penalty0 1--10, 2018.

\bibitem[Xue et~al.(2017)Xue, Zhang, Deng, Chen, and Jiang]{xue2017preliminary}
Yanping Xue, Rongguo Zhang, Yufeng Deng, Kuan Chen, and Tao Jiang.
\newblock A preliminary examination of the diagnostic value of deep learning in
  hip osteoarthritis.
\newblock \emph{PLoS One}, 12\penalty0 (6):\penalty0 e0178992, 2017.

\bibitem[Simonyan and Zisserman(2014)]{simonyan2014very}
Karen Simonyan and Andrew Zisserman.
\newblock Very deep convolutional networks for large-scale image recognition.
\newblock \emph{arXiv preprint arXiv:1409.1556}, 2014.

\bibitem[Rohrbach et~al.(2019)Rohrbach, Reinhard, Sick, and
  D{\"u}rr]{rohrbach2019bone}
Janick Rohrbach, Tobias Reinhard, Beate Sick, and Oliver D{\"u}rr.
\newblock Bone erosion scoring for rheumatoid arthritis with deep convolutional
  neural networks.
\newblock \emph{Computers \& Electrical Engineering}, 78:\penalty0 472--481,
  2019.

\bibitem[Hirano et~al.(2019)Hirano, Nishide, Nonaka, Seita, Ebina, Sakurada,
  and Kumanogoh]{hirano2019development}
Toru Hirano, Masayuki Nishide, Naoki Nonaka, Jun Seita, Kosuke Ebina, Kazuhiro
  Sakurada, and Atsushi Kumanogoh.
\newblock Development and validation of a deep-learning model for scoring of
  radiographic finger joint destruction in rheumatoid arthritis.
\newblock \emph{Rheumatology advances in practice}, 3\penalty0 (2):\penalty0
  rkz047, 2019.

\bibitem[Lin et~al.(2017{\natexlab{a}})Lin, Goyal, Girshick, He, and
  Doll{\'a}r]{lin2017focal}
Tsung-Yi Lin, Priya Goyal, Ross Girshick, Kaiming He, and Piotr Doll{\'a}r.
\newblock Focal loss for dense object detection.
\newblock In \emph{Proceedings of the IEEE international conference on computer
  vision}, pages 2980--2988, 2017{\natexlab{a}}.

\bibitem[He et~al.(2016)He, Zhang, Ren, and Sun]{he2016deep}
Kaiming He, Xiangyu Zhang, Shaoqing Ren, and Jian Sun.
\newblock Deep residual learning for image recognition.
\newblock In \emph{Proceedings of the IEEE conference on computer vision and
  pattern recognition}, pages 770--778, 2016.

\bibitem[Huang et~al.(2017)Huang, Liu, Van Der~Maaten, and
  Weinberger]{huang2017densely}
Gao Huang, Zhuang Liu, Laurens Van Der~Maaten, and Kilian~Q Weinberger.
\newblock Densely connected convolutional networks.
\newblock In \emph{Proceedings of the IEEE conference on computer vision and
  pattern recognition}, pages 4700--4708, 2017.

\bibitem[Lin et~al.(2017{\natexlab{b}})Lin, Doll{\'a}r, Girshick, He,
  Hariharan, and Belongie]{lin2017feature}
Tsung-Yi Lin, Piotr Doll{\'a}r, Ross Girshick, Kaiming He, Bharath Hariharan,
  and Serge Belongie.
\newblock Feature pyramid networks for object detection.
\newblock In \emph{Proceedings of the IEEE conference on computer vision and
  pattern recognition}, pages 2117--2125, 2017{\natexlab{b}}.

\bibitem[Li and Ren(2019)]{li2019light}
Yixing Li and Fengbo Ren.
\newblock Light-weight retinanet for object detection.
\newblock \emph{arXiv preprint arXiv:1905.10011}, 2019.

\bibitem[Lin et~al.(2014)Lin, Maire, Belongie, Hays, Perona, Ramanan,
  Doll{\'a}r, and Zitnick]{lin2014microsoft}
Tsung-Yi Lin, Michael Maire, Serge Belongie, James Hays, Pietro Perona, Deva
  Ramanan, Piotr Doll{\'a}r, and C~Lawrence Zitnick.
\newblock Microsoft coco: Common objects in context.
\newblock In \emph{European conference on computer vision}, pages 740--755.
  Springer, 2014.

\bibitem[Wang et~al.(2019)Wang, Wang, Zhang, Dong, and Wei]{wang2019automatic}
Yuanyuan Wang, Chao Wang, Hong Zhang, Yingbo Dong, and Sisi Wei.
\newblock Automatic ship detection based on retinanet using multi-resolution
  gaofen-3 imagery.
\newblock \emph{Remote Sensing}, 11\penalty0 (5):\penalty0 531, 2019.

\bibitem[Tan and Le(2019)]{tan2019efficientnet}
Mingxing Tan and Quoc~V Le.
\newblock Efficientnet: Rethinking model scaling for convolutional neural
  networks.
\newblock \emph{arXiv preprint arXiv:1905.11946}, 2019.

\bibitem[Szegedy et~al.(2015)Szegedy, Liu, Jia, Sermanet, Reed, Anguelov,
  Erhan, Vanhoucke, and Rabinovich]{szegedy2015going}
Christian Szegedy, Wei Liu, Yangqing Jia, Pierre Sermanet, Scott Reed, Dragomir
  Anguelov, Dumitru Erhan, Vincent Vanhoucke, and Andrew Rabinovich.
\newblock Going deeper with convolutions.
\newblock In \emph{Proceedings of the IEEE conference on computer vision and
  pattern recognition}, pages 1--9, 2015.

\bibitem[Huang et~al.(2019)Huang, Cheng, Bapna, Firat, Chen, Chen, Lee, Ngiam,
  Le, Wu, et~al.]{huang2019gpipe}
Yanping Huang, Youlong Cheng, Ankur Bapna, Orhan Firat, Dehao Chen, Mia Chen,
  HyoukJoong Lee, Jiquan Ngiam, Quoc~V Le, Yonghui Wu, et~al.
\newblock Gpipe: Efficient training of giant neural networks using pipeline
  parallelism.
\newblock In \emph{Advances in neural information processing systems}, pages
  103--112, 2019.

\bibitem[Xu et~al.(2015)Xu, Ba, Kiros, Cho, Courville, Salakhudinov, Zemel, and
  Bengio]{xu2015show}
Kelvin Xu, Jimmy Ba, Ryan Kiros, Kyunghyun Cho, Aaron Courville, Ruslan
  Salakhudinov, Rich Zemel, and Yoshua Bengio.
\newblock Show, attend and tell: Neural image caption generation with visual
  attention.
\newblock In \emph{International conference on machine learning}, pages
  2048--2057, 2015.

\bibitem[Schlemper et~al.(2019)Schlemper, Oktay, Schaap, Heinrich, Kainz,
  Glocker, and Rueckert]{schlemper2019attention}
Jo~Schlemper, Ozan Oktay, Michiel Schaap, Mattias Heinrich, Bernhard Kainz, Ben
  Glocker, and Daniel Rueckert.
\newblock Attention gated networks: Learning to leverage salient regions in
  medical images.
\newblock \emph{Medical image analysis}, 53:\penalty0 197--207, 2019.

\bibitem[Bridges~Jr et~al.(2010)Bridges~Jr, Causey, Burgos, Huynh, Hughes,
  Danila, Van~Everdingen, Ledbetter, Conn, Tamhane,
  et~al.]{bridges2010radiographic}
S~Louis Bridges~Jr, Zenoria~L Causey, Paula~I Burgos, B~Quynh~N Huynh, Laura~B
  Hughes, Maria~I Danila, Amalia Van~Everdingen, Stephanie Ledbetter, Doyt~L
  Conn, Ashutosh Tamhane, et~al.
\newblock Radiographic severity of rheumatoid arthritis in african americans:
  results from a multicenter observational study.
\newblock \emph{Arthritis care \& research}, 62\penalty0 (5):\penalty0
  624--631, 2010.

\bibitem[Ormseth et~al.(2016)Ormseth, Yancey, Solus, Bridges~Jr, Curtis,
  Linton, Fazio, Davies, Roberts, Vickers, et~al.]{ormseth2016effect}
Michelle~J Ormseth, Patricia~G Yancey, Joseph~F Solus, S~Louis Bridges~Jr,
  Jeffrey~R Curtis, MacRae~F Linton, Sergio Fazio, Sean~S Davies, L~Jackson
  Roberts, Kasey~C Vickers, et~al.
\newblock Effect of drug therapy on net cholesterol efflux capacity of
  high-density lipoprotein--enriched serum in rheumatoid arthritis.
\newblock \emph{Arthritis \& rheumatology}, 68\penalty0 (9):\penalty0
  2099--2105, 2016.

\bibitem[Sun et~al.(2021)Sun, Nguyen, Allaway, Wang, Chung, Yu, Mason,
  Dimitrovsky, Ericson, Li, Guan, Israel, Olar, Pataki, Community, Stolovitzky,
  Guinney, Gulko, Frazier, Costello, Chen, and Bridges]{Sun2021.10.25.21265495}
Dongmei Sun, Thanh~M. Nguyen, Robert~J. Allaway, Jelai Wang, Verena Chung,
  Thomas~V Yu, Michael Mason, Isaac Dimitrovsky, Lars Ericson, Hongyang Li,
  Yuanfang Guan, Ariel Israel, Alex Olar, Balint~Armin Pataki, RA2
  DREAM~Challenge Community, Gustavo Stolovitzky, Justin Guinney, Percio~S.
  Gulko, Mason~B. Frazier, James~C. Costello, Jake~Y. Chen, and S.~Louis
  Bridges.
\newblock A crowdsourcing approach to develop machine learning models to
  quantify radiographic joint damage in rheumatoid arthritis.
\newblock \emph{medRxiv}, 2021.
\newblock \doi{10.1101/2021.10.25.21265495}.
\newblock URL
  \url{https://www.medrxiv.org/content/early/2021/10/26/2021.10.25.21265495}.

\bibitem[Kingma and Ba(2014)]{kingma2014adam}
Diederik~P Kingma and Jimmy Ba.
\newblock Adam: A method for stochastic optimization.
\newblock \emph{arXiv preprint arXiv:1412.6980}, 2014.

\end{thebibliography}

\end{document}